%% file: main.tex
\documentclass{article}
\usepackage[accepted]{icml2018}

\usepackage{times}
\usepackage{graphicx} 

\usepackage{algorithm}
\usepackage{algorithmic}
\usepackage{arydshln}

\PassOptionsToPackage{numbers, compress}{natbib}
\usepackage{natbib}

\usepackage{booktabs}
\usepackage{color}
\usepackage{placeins}
\usepackage{amsmath}
\usepackage{amsthm}
\usepackage{amssymb}
\usepackage{caption}
\usepackage{subcaption}
\usepackage{tikz}
\usepackage{comment}
\usepackage{bm}
\usepackage{scalefnt}

\newtheorem{prop}{Proposition}

\newcommand\numberthis{\refstepcounter{equation}\tag{\theequation}}

\newcommand{\cv}{\boldsymbol{c}}

\newcommand{\xv}{\boldsymbol{x}}

\newcommand{\zv}{\boldsymbol{z}}

\newcommand{\Eb}{\mathbb{E}}

\newcommand{\cov}{\mathrm{Cov}}
\newcommand{\data}{\mathcal{D}}
\newcommand{\KL}{D_{\mathrm{KL}}}

\newcommand{\MMD}{D_{\mathrm{MMD}}}
\newcommand{\InfoVAE}{\mathcal{L}_{\mathrm{InfoVAE}}}
\newcommand{\InfoVAEh}{\hat{\mathcal{L}}_{\mathrm{InfoVAE}}}
\newcommand{\ELBO}{\mathcal{L}_{\mathrm{ELBO}}}

\def\E{\mathbb{E}}

\begin{document}

\twocolumn[
\icmltitle{InfoVAE: Balancing Learning and Inference in Variational Autoencoders}



\icmlsetsymbol{equal}{*}

\begin{icmlauthorlist}
\icmlauthor{Shengjia Zhao}{stanford}
\icmlauthor{Jiaming Song}{stanford}
\icmlauthor{Stefano Ermon}{stanford}
\end{icmlauthorlist}

\icmlaffiliation{stanford}{Computer Science Department, Stanford University}

\icmlcorrespondingauthor{Shengjia Zhao}{sjzhao@stanford.edu}
\icmlcorrespondingauthor{Stefano Ermon}{ermon@stanford.edu}


\icmlkeywords{Machine Learning, ICML}

\vskip 0.3in
]

\printAffiliationsAndNotice{}  

\begin{abstract}
A key advance in learning generative models is the use of amortized inference distributions that are jointly trained with the models. We find that existing training objectives for variational autoencoders can lead to inaccurate amortized inference distributions and, in some cases, improving the objective provably degrades the inference quality. In addition, it has been observed that variational autoencoders tend to ignore the latent variables when combined with a decoding distribution that is too flexible. We again identify the cause in existing training criteria and propose a new class of objectives (InfoVAE) that mitigate these problems. We show that our model can significantly improve the quality of the variational posterior and can make effective use of the latent features regardless of the flexibility of the decoding distribution. Through extensive qualitative and quantitative analyses, we demonstrate that our models outperform competing approaches on multiple performance metrics. 
\end{abstract}
 
 
\section{Introduction}
Generative models have shown great promise in modeling complex distributions such as natural images and text~\citep{deconvolutional_gan2015,cyclegan2017,improved_vae_nlp2017,infogail2017}. 
These are directed graphical models
which represent the joint distribution between the data and
a set of hidden variables (features) capturing latent factors
of variation. The joint is factored as the product of a prior
over the latent variables and a conditional distribution of
the visible variables given the latent ones.
Usually a simple prior distribution is provided for the latent variables, while the distribution of the input conditioned on latent variables is complex and modeled with a deep network. 

Both learning and inference are generally intractable. However, using an amortized approximate inference distribution it is possible to use the evidence lower bound (ELBO) to efficiently optimize 
both (a lower bound to) the marginal likelihood of the data and the quality of the approximate inference distribution.
This leads to a very successful class of models called variational autoencoders~\citep{autoencoding_variational_bayes2013, variational_dbn_stochastic_bp2014,vae_autoregressive_flow2016,importance_weighted_autoencoders2015}.

However, variational autoencoders have several problems. First, the approximate inference distribution is often significantly different from the true posterior. Previous methods have resorted to using more flexible variational families to better approximate the true posterior distribution~\citep{vae_autoregressive_flow2016}. However we find that
the problem is rooted in the ELBO training objective itself. In fact, we show that the 
ELBO objective favors fitting 
the data distribution over performing correct amortized inference. When the two goals are conflicting (e.g., because of limited capacity), the  ELBO objective tends to sacrifice correct inference to better fit (or worse overfit) the training data. 


Another problem that has been observed is that when the conditional distribution is sufficiently expressive, the latent variables are often ignored~\citep{lossy_vae2016}. That is, the model only uses a single conditional distribution component to model the data, effectively ignoring the latent variables and fail to take advantage of the mixture modeling capability of the VAE. In addition, one goal of unsupervised learning is to learn meaningful latent representations but this fails because the latent variables are ignored. Some solutions have been proposed in~\citep{lossy_vae2016} by limiting the capacity of the conditional distribution, but this requires manual and problem-specific design of the features we would like to extract. 

In this paper we propose a novel solution by framing both problems as explicit modeling choices: we introduce new training objectives where it is possible to weight the preference 
between 
correct inference and fitting data distribution, and specify a preference on how much the model should rely on the latent variables.
This choice is implicitly made in the ELBO objective. We make this choice explicit and generalize the ELBO objective by adding additional terms that allow users to select their preference on both choices.
Despite of the addition of seemingly intractable terms, we find an equivalent form that can still be efficiently optimized. 

Our new family also generalizes known models including the $\beta$-VAE~\citep{beta_vae2016} and Adversarial Autoencoders~\citep{adversarial_autoencoder2015}. In addition to deriving these models as special cases, 
we provide generic principles for hyper-parameter selection that work well in all the experimental settings we considered. 
Finally we perform extensive experiments to evaluate our newly introduced model family, and compare 
with
existing models on multiple metrics of performance such as log-likelihood, sampling quality, and semi-supervised performance. An instantiation of our general framework called MMD-VAE achieves better or on-par performance on all metrics we considered. 
We further observe that our model can lead to better amortized inference, and utilize the latent variables even in the presence of a very flexible decoder. 


\section{Variational Autoencoders}
\label{sec:information_preference}
A latent variable generative model defines a joint distribution between a feature space $z \in \mathcal{Z}$, and the input space $x \in \mathcal{X}$. Usually we assume a simple prior distribution $p(z)$ over features, such as Gaussian or uniform, and model the data distribution with a complex conditional distribution $p_\theta(x|z)$, where $p_\theta(x|z)$ is often parameterized with a neural network. Suppose the true underlying distribution is $p_\data(x)$ (that is approximated by a training set), then a natural training objective is maximum (marginal) likelihood
\begin{equation}
\label{eq:ELBO}
\Eb_{p_{\data}(x)}[\log p_\theta(x)] = \Eb_{p_{\data}(x)}[\log \Eb_{p(z)}[p_\theta(x|z)]] \end{equation}
However direct optimization of the likelihood is intractable because computing $p_\theta(x) = \int_z p_\theta(x|z) p(z) \mathrm{d}z$ requires integration. A classic approach \citep{autoencoding_variational_bayes2013} is to define an amortized inference distribution $q_\phi(z|x)$ and jointly optimize a lower bound to the log likelihood
\begin{align*}
\ELBO(x) &= -\KL(q_\phi(z|x)||p(z)) + \Eb_{q_\phi(z|x)}[\log p_\theta(x|z)] \\ 
&\leq \log p_\theta(x) 
\end{align*}

We further average this over the data distribution $p_\data(x)$ to obtain the final optimization objective
\[ \ELBO = \Eb_{p_\data(x)}[\ELBO(x)] \leq \Eb_{p_\data(x)}[\log p_\theta(x)] \]

\subsection{Equivalent Forms of the ELBO Objective}
There are several ways to equivalently rewrite the ELBO objective that will become useful in our following analysis. We define the joint generative distribution as 
\[ p_\theta(x, z) \equiv p(z) p_\theta(x|z) \]
In fact we can correspondingly define a joint ``inference distribution''
\[ q_\phi(x, z) \equiv p_\data(x) q_\phi(z|x) \]
Note that the two definitions are symmetrical. In the former case we start from a known distribution $p(z)$ and learn the conditional distribution on $\mathcal{X}$, in the latter we start from a known (empirical) distribution $p_\data(x)$ and learn the conditional distribution on $\mathcal{Z}$. We also correspondingly define any conditional and marginal distributions as follows: 
\begin{align*}
p_\theta(x) = \int_z p_\theta(x, z) dz & \qquad \mathrm{Marginal\ of\ } p_\theta(x, z) \mathrm{\ on\ } x \\ 
p_\theta(z|x) \propto p_\theta(x, z) & \qquad \mathrm{Posterior\ of\ } p_\theta(x|z) \\
q_\phi(z) = \int_x q_\phi(x, z) dx & \qquad \mathrm{Marginal\ of\ } q_\phi(x, z) \mathrm{\ on\ } z \\ 
q_\phi(x|z) \propto q_\phi(x, z) & \qquad \mathrm{Posterior\ of\ }q_\phi(z|x)
\end{align*}
For the purposes of optimization, the ELBO objective can be written equivalently (up to an additive constant) as 
\begin{align*}
\ELBO &\equiv -\KL(q_\phi(x, z) \Vert p_\theta(x, z)) \numberthis \label{eq:elbo_form1} \\ 
&= -\KL(p_\data(x) \Vert p_\theta(x)) \numberthis \label{eq:elbo_form2} \\
& \qquad \qquad -\Eb_{p_\data(x)}[\KL(q_\phi(z|x) \Vert p_\theta(z|x)) ]  \\
&= -\KL(q_\phi(z) \Vert p(z))  \\
& \qquad \qquad -\Eb_{q_\phi(z)}[\KL(q_\phi(x|z) \Vert p_\theta(x|z)) ]  \numberthis \label{eq:elbo_form3}
\end{align*}
We prove the first equivalence in the appendix. The second and third equivalence are simple applications of the additive property of KL divergence. All three forms of ELBO in Eqns. (\ref{eq:elbo_form1}),(\ref{eq:elbo_form2}),(\ref{eq:elbo_form3}) are useful in our analysis. 


\section{Two Problems of Variational Autoencoders}
\subsection{Amortized Inference Failures}
Under ideal conditions, optimizing the ELBO objective using sufficiently flexible model families for $p_\theta(x|z)$ and $q_\phi(z|x)$ over $\theta,\phi$ will achieve both goals of correctly capturing $p_\data(x)$ and performing correct amortized inference. 
This can be seen by examining Eq. (\ref{eq:elbo_form2}).
This form indicates that the ELBO objective is minimizing the KL divergence between the data distribution $p_\data(x)$ and the (marginal) model distribution $p_\theta(x)$, as well as the KL divergence between the variational posterior $q_\phi(z|x)$ and the true posterior $p_\theta(z|x)$.
However, with finite model capacity the two goals can be conflicting and subtle tradeoffs and failure modes can emerge from optimizing the ELBO objective. 

In particular, one limitation of the ELBO objective is that it might fail to learn an amortized inference distribution $q_\phi(z|x)$ that approximates the true posterior $p_\theta(z|x)$. This can happen for two different reasons: 

\textbf{Inherent properties of the ELBO objective:} the ELBO objective can be maximized (even to $+\infty$ in pathological cases) even with a very inaccurate variational posterior $q_\phi(z|x)$. 

\textbf{Implicit modeling bias:} common modeling choices (such as the high dimensionality of $\mathcal{X}$ compared to $\mathcal{Z}$) tend to sacrifice variational inference vs. data fit when modeling capacity is not sufficient to achieve both. 

We will explain in turn why these failures happen.

\subsubsection{Good ELBO Values Do Not Imply Accurate Inference}
\label{sec:inference}
We first provide some intuition to this phenomena, then formally prove the result for a pathological case of continuous spaces and Gaussian distributions. Finally we justify in the experiments section that this happens in realistic settings on real datasets (in both continuous and discrete spaces).

The ELBO objective in original form has two components, a log likelihood (reconstruction) term $\mathcal{L}_{\mathrm{AE}}$ and a regularization term $\mathcal{L}_{\mathrm{REG}}$:
\begin{align*}
&\ELBO(x) = \mathcal{L}_{\mathrm{AE}}(x) +  \mathcal{L}_{\mathrm{REG}}(x) \\
& \equiv \Eb_{q_\phi(z|x)}[\log p_\theta(x|z)] - \KL(q_\phi(z|x)||p(z))
\end{align*}
Let us first consider what happens if we only optimize $ \mathcal{L}_{\mathrm{AE}}$ and not $\mathcal{L}_{\mathrm{REG}}$. The first term maximizes the log likelihood of observing data point $x$ given its inferred latent variables $z \sim q_\phi(z|x)$. Consider a finite dataset $\lbrace x_1, \cdots, x_N \rbrace$. Let $q_\phi$ be such that for $x_i \neq x_j$, $q_\phi(z|x_i)$ and $q_\phi(z|x_j)$ are distributions with disjoint supports. Then we can learn a $p_\theta(x|z)$ mapping the support of each $q_\phi(z|x_i)$ to a distribution concentrated on $x_i$, leading to very large $\mathcal{L}_{\mathrm{AE}}$ (for continuous distributions $p_\theta(x|z)$ may even tend to a Dirac delta distribution and $\mathcal{L}_{\mathrm{AE}}$ tends to $+\infty$). Intuitively, the $\mathcal{L}_{\mathrm{AE}}$ component will encourage choosing $q_\phi(z|x_i)$ with disjoint support when $x_i \neq x_j$.  

In almost all practical cases, the variational distribution family for $q_\phi$ is supported on the entire space $\mathcal{Z}$ (e.g., it is a Gaussian with non-zero variance, or IAF posterior~\citep{vae_autoregressive_flow2016}), preventing disjoint supports.
However, attempting to learn disjoint supports for $q_\phi(z|x_i), x_i \neq x_j$ will "push" the mass of the distributions away from each other. For example, for continuous distributions, if $q_\phi$ maps each $x_i$ to a Gaussian $\mathcal{N}(\mu_i, \sigma_i)$, the $\mathcal{L}_{\mathrm{AE}}$ term will encourage $\mu_i \to \infty, \sigma_i \to 0^+$. 

This undesirable result may be prevented if the $\mathcal{L}_{\mathrm{REG}}$ term can counter-balance this tendency. However, we show that the regularization term $\mathcal{L}_{\mathrm{REG}}$ is not always sufficiently strong to prevent this issue.
We first prove this fact in the simple case of a mixture of two Gaussians. We will then evaluate this finding empirically on realistic datasets in the experiments section \ref{experiments:exploding}.


\begin{prop}
\label{prop:mixture_of_gaussian}
Let $\mathcal{D}$ be a dataset with two samples $\lbrace -1, 1 \rbrace$, and $p_\theta(x|z)$ be selected from the family of all functions $\mu^p_\theta, \sigma^p_\theta$ that map $z\in\mathcal{X}$ to a Gaussian $\mathcal{N}(\mu^p_\theta(z), \sigma^p_\theta(z))$ on $\mathcal{X}$, and $q_\phi(z|x)$ be selected from the family of all functions $\mu^q_\phi, \sigma^q_\phi$ that map $x\in\mathcal{X}$ to a Gaussian $\mathcal{N}(\mu^q_\phi(z), \sigma^q_\phi(z))$ on $\mathcal{Z}$. Then $\mathcal{L}_{ELBO}$ can be maximized to $+\infty$ when
\begin{align*}
&\mu^q_\phi(x=1) \to +\infty \qquad & \mu^q_\phi(x=-1) &\to -\infty \\
&\sigma^q_\phi(x=1) \to 0^+  \qquad & \sigma^q_\phi(x=-1) &\to 0^+
\end{align*}
and $\theta$ is optimally selected given $\phi$. In addition the variational gap $\KL(q_\phi(z|x)\Vert p_\theta(z|x)) \to +\infty$ for all $x \in \mathcal{D}$.
\end{prop}

A proof can be found the in Appendix. This means that amortized inference has completely failed, even though the ELBO objective can be made arbitrarily large. The model learns an inference distribution $q_\phi(z|x)$ that pushes all probability mass to $\infty$. This will become infinitely far from the true posterior $p_\theta(z|x)$ as measured by $\KL$. 

\subsubsection{Modeling Bias}
In the above example we indicated a potential problem with the ELBO objective where the model tends to push the probability mass of $q_\phi(z|x)$ too far from $0$. This tendency is a property of the ELBO objective and true for any $\mathcal{X}$ and $\mathcal{Z}$. However this is made worse by the fact that $\mathcal{X}$ is often higher dimensional compared to $\mathcal{Z}$, so any error in fitting $\mathcal{X}$ will be magnified compared to $\mathcal{Z}$.  

For example, consider fitting an $n$ dimensional distribution $\mathcal{N}(0, I)$ with $\mathcal{N}(\epsilon, I)$ using KL divergence, then 
\[ \KL(\mathcal{N}(0, I), \mathcal{N}(\epsilon, I)) = n \epsilon^2/2 \]
As $n$ increases with some fixed $\epsilon$, the Euclidean distance between the means of the two distributions is $\Theta(\sqrt{n})$, yet the corresponding $\KL$ becomes $\Theta(n)$.
For natural images, the dimensionality of $\mathcal{X}$ is often orders of magnitude larger than the dimensionality of $\mathcal{Z}$. Recall in Eq.(\ref{eq:elbo_form3}) that ELBO is optimizing both $\KL(q_\phi(z)\Vert p(z))$ and $\KL(q_\phi(x|z)\Vert p_\theta(x|z))$. Because the same per-dimensional modeling error incurs a much larger loss in $\mathcal{X}$ space than $\mathcal{Z}$ space, when the two objectives are conflicting (e.g., because of limited modeling capacity), the model will tend to sacrifice divergences on $\mathcal{Z}$ and focus on minimizing divergences on $\mathcal{X}$. 

Regardless of the cause (properties of ELBO or modeling choices), this is generally an undesirable phenomenon for two reasons: 

\textbf{1)} One may care about accurate inference more than generating sharp samples. For example, generative models are often used for down stream tasks such as semi supervised learning. 

\textbf{2)} Overfitting: Because $p_\data$ is an empirical (finite) distribution in practice, matching it too closely can lead to poor generalization. 


Both issues are observed in the experiments section.

\subsection{The Information Preference Property}
\label{sec:ipp}
Using a complex decoding distribution $p_\theta(x|z)$ such as PixelRNN/PixelCNN~\citep{pixel_rnn2015,pixel_vae2016} has been shown to significantly improve sample quality on complex natural image datasets. However, this approach suffers from a new problem: it tends to neglect the latent variables $z$ altogether, that is, the mutual information 
between $z$ and $x$ becomes vanishingly small. Intuitively, the reason is that the learned $p_\theta(x|z)$ is the same for all $z \in \mathcal{Z}$, implying that the $z$ is not dependent on the input $x$. This is undesirable because a major goal of unsupervised learning is to learn meaningful latent features which \emph{should} depend on the inputs.

This effect, which we shall refer to as the \emph{information preference} problem, was studied in \citep{lossy_vae2016} with a coding efficiency argument. Here we provide an alternative interpretation, which sheds light on a novel solution to solve this problem.

We inspect the ELBO in the form of Eq.(\ref{eq:elbo_form2}), and consider the two terms respectively. We show that both can be optimized to $0$ without utilizing the latent variables. 

\textbf{$\KL(p_\data(x)||p_\theta(x))$:} Suppose the model family $\{p_\theta(\cdot|z),\theta \in \Theta\}$ is sufficiently flexible and there exists a $\theta^*$ such that for every $z\in \mathcal{Z}$, $p_{\theta^*}(\cdot|z)$ and $ p_\data(\cdot)$ are identical. Then we select this $\theta^*$ and the marginal $p_{\theta^*}(x) = \int_z p(z) p_{\theta^*}(x|z) dz 
= p_\data(x)$, hence this divergence $\KL(p_\data(x)||p_\theta(x)) = 0$ which is optimal. 

\textbf{$\Eb_{p_\data(x)}[\KL(q_\phi(z|x)||p_\theta(z|x))]$:} Because $p_{\theta^*}(\cdot|z)$ is the same for every $z$ ($x$ is independent from $z$) we have $p_{\theta^*}(z|x) = p(z)$. Because $p(z)$ is usually a simple distribution, if it is possible to choose $\phi$ such that $q_\phi(z|x) = p(z), \forall x \in \mathcal{X}$, this divergence will also achieve the optimal value of $0$. 

Because $\mathcal{L}_{\mathrm{ELBO}}$ is the sum of the above divergences, when both are $0$, this is a global optimum. There is no incentive for the model to learn otherwise, undermining our purpose of learning a latent variable model. 

\section{The InfoVAE Model Family}
%
%

In order to remedy these two problems we define a new training objective that will learn both the correct model and amortized inference distributions. We begin with the form of ELBO in Eq. (\ref{eq:elbo_form3}) 
\begin{align*}
\ELBO &= -\KL(q_\phi(z) \Vert p(z)) - \\
& \quad \Eb_{p(z)}[\KL(q_\phi(x|z) \Vert p_\theta(x|z)) ] 
\end{align*}

First we add a scaling parameter $\lambda$ to the divergence between $q_\phi(z)$ and $p(z)$ to increase its weight and counter-act the imbalance between $\mathcal{X}$ and $\mathcal{Z}$ (cf. discussion in section \ref{sec:inference}). Next we add a mutual information maximization term that prefers high mutual information between $x$ and $z$. This encourages the model to use the latent code and avoids the information preference problem. 
We arrive at the following objective
\begin{align*}
& \InfoVAE = -\lambda \KL(q_\phi(z) \Vert p(z)) - \\
& \qquad \Eb_{q(z)}[\KL(q_\phi(x|z) \Vert p_\theta(x|z)) ] + \alpha I_q(x; z) \numberthis \label{eq:info_vae}
\end{align*}
where $I_q(x; z)$ is the mutual information between $x$ and $z$ under the distribution $q_\phi(x, z)$. 

Even though we cannot directly optimize this objective, we can rewrite this into an equivalent form that we can optimize more efficiently (We prove this in the Appendix)
\begin{align*}
\InfoVAE &\equiv \Eb_{p_\data(x)} \Eb_{q_\phi(z|x)}[\log p_\theta(x|z)] -\\ 
& \quad (1 - \alpha) \Eb_{p_\data(x)} \KL(q_\phi(z|x)||p(z)) - \\
& \quad (\alpha + \lambda - 1) \KL(q_\phi(z) \Vert p(z)) \numberthis \label{eq:info_vae_tract}
\end{align*}
The first two terms can be optimized by the reparameterization trick as in the original ELBO objective. The last term $\KL(q_\phi(z) \Vert p(z))$ is not easy to compute because we cannot tractably evaluate $\log q_\phi(z)$. However we can obtain unbiased samples from it by first sampling $x \sim p_\data$, then $z \sim q_\phi(z|x)$, so we can optimize it by likelihood free optimization techniques~\citep{generative_adversarial_nets2014,f_gan2016,wgan2017,mmd_statistics2007}. In fact we may replace the term $\KL(q_\phi(z) \Vert p(z))$ with anther divergence $D(q_\phi(z) \Vert p(z))$ that we can efficiently optimize over. Changing the divergence may alter the empirical behavior of the model but we show in the following theorem that replacing $\KL$ with any (strict) divergence is still correct. Let $\hat{\mathcal{L}}_{\mathrm{InfoVAE}}$ be the objective where we replace $\KL(q_\phi(z) \Vert p(z))$ with any strict divergence $D(q_\phi(z) \Vert p(z))$. (Strict divergence is defined as $D(q_\phi(z) \Vert p(z))=0$ iff $q_\phi(z)=p(z)$) 
\begin{prop}
\label{prop:infovae_correctness}
Let $\mathcal{X}$ and $\mathcal{Z}$ be continuous spaces, and $\alpha < 1$, $\lambda > 0$. For any fixed value of $I_q(x; z)$, $\hat{\mathcal{L}}_{\mathrm{InfoVAE}}$ is globally optimized 
if
$p_\theta(x)=p_{\data}(x)$ and $q_\phi(z|x) =p_\theta(z|x), \forall z$.
\end{prop}
\begin{proof}[Proof of Proposition~\ref{prop:infovae_correctness}]
See Appendix.
\end{proof}
Note that in the proposition we have the additional requirement that the mutual information $I_q(x; z)$ is bounded. This is inevitable because if $\alpha > 0$ the objective can be optimized to $+\infty$ by simply increasing the mutual information infinitely. In our experiments simply ensuring that $q_\phi(z|x)$ does not have vanishing variance is sufficient to regularize the behavior of the model.

\textbf{Relation to VAE and $\beta$-VAE:} This model family generalizes several previous models. When $\alpha=0$ and $\lambda=1$ we get back the original ELBO objective. When $\lambda > 0$ is freely chosen, while $\alpha + \lambda - 1 = 0$, and we use the $\KL$ divergences, we get the $\beta$-VAE~\citep{beta_vae2016} model family. However, $\beta$-VAE models cannot effectively trade-off weighing of $\mathcal{X}$ and $\mathcal{Z}$ and information preference. In particular, 
for every $\lambda$ there is a unique value of $\alpha$ that we can choose. For example, if we choose a large value of $\lambda \gg 1$ to balance the importance of observed and latent spaces $\mathcal{X}$ and $\mathcal{Z}$, we must also choose $\alpha \ll 0$, which forces the model to penalize mutual information. This in turn can lead to under-fitting or ignoring the latent variables. 

\textbf{Relation to Adversarial Autoencoders (AAE):} When $\alpha=1$, $\lambda=1$ and $D$ is chosen to be the Jensen Shannon divergence we get the adversarial autoencoders in \citep{adversarial_autoencoder2015}. This paper generalizes AAE, but more importantly we provide a deeper understanding of the correctness and desirable properties of AAE. Furthermore, we characterize settings when AAE is preferable compared to VAE (i.e. when we would like to have $\alpha=1$). 

Our generalization introduces new parameters, but the meaning and effect of the various parameter choices is clear. We recommend setting $\lambda$ to a value that makes the loss on $\mathcal{X}$ approximately the same as the loss on $\mathcal{Z}$. We also recommend setting $\alpha = 0$ when $p_\theta(x|z)$ is a simple distribution, and $\alpha=1$ when $p_\theta(x|z)$ is a complex distribution and information preference is a concern. The final degree of freedom is the divergence $D(q_\phi(z)\Vert p(z))$ to use. We will explore this topic in the next section. We will also show in the experiments that this design choice is also easy to choose: there is a choice that we find to be consistently better in almost all metrics of performance.

\subsection{Divergences Families}
We consider and compare three divergences in this paper. 

\textbf{Adversarial Training:} Adversarial autoencoders (AAE) proposed in \citep{adversarial_autoencoder2015} use an adversarial discriminator to approximately minimize the Jensen-Shannon divergence \citep{generative_adversarial_nets2014} between $q_\phi(z)$ and $p(z)$. However, when $p(z)$ is a simple distribution such as Gaussian, there are preferable alternatives. In fact, adversarial training can be unstable and slow even when we apply recent techniques for stabilizing GAN training \citep{wgan2017,improved_wgan2017}.

\textbf{Stein Variational Gradient:} The Stein variational gradient \citep{stein_variational2016} is a simple and effective framework for matching a distribution $q$ to $p$ by computing effectively $ \nabla_{\phi} \KL(q_\phi(z) || p(z)) $
which we can use for gradient descent minimization of $\KL(q_\phi(z) || p(z))$. However a weakness of these methods is that they are difficult to apply efficiently in high dimensions. We give a detailed overview of this method in the Appendix.  



\textbf{Maximum-Mean Discrepancy:} Maximum-Mean Discrepancy (MMD) \citep{mmd_statistics2007,generative_moment_matching_network2014,mmd_network2015} is a framework to quantify the distance between two distributions by comparing all of their moments. It can be efficiently implemented using the kernel trick. Letting $k(\cdot,\cdot)$ be any positive definite kernel, the MMD between $p$ and $q$ is
\begin{align*}
\MMD(q \Vert p) &= \Eb_{p(z), p(z')}[k(z, z')] - 2 \Eb_{q(z), p(z')}[k(z, z')] \\ &+ \Eb_{q(z), q(z')}[k(z, z')]
\end{align*}
$\MMD = 0$ if and only if $p = q$.




\section{Experiments}
\input{experiments}
\section{Conclusion}
Despite the recent success of variational autoencoders, they can fail to perform amortized inference, or learn meaningful latent features. We trace both issues back to the ELBO learning criterion, and modify the ELBO objective to propose a new model family that can fix both problems. We perform extensive experiments to verify the effectiveness of our approach. Our experiments show that a particular subset of our model family, MMD-VAEs perform on-par or better than all other approaches on multiple metrics of performance. 

\section{Acknowledgements}
We thank Daniel Levy, Rui Shu, Neal Jean, Maria Skoularidou for providing constructive feedback and discussions.

\FloatBarrier


\bibliographystyle{icml2018}
\begin{small}
\bibliography{ref}
\end{small}

\clearpage
\newpage
\FloatBarrier
\input{appendix}

\end{document}

%% file: experiments.tex
\subsection{Variance Overestimation with ELBO training}
\label{experiments:exploding}
We first perform some simple experiments on toy data and MNIST to demonstrate that ELBO suffers from inaccurate inference in practice, and adding the scaling term $\lambda$ in Eq.(\ref{eq:info_vae}) can correct for this. Next, we will perform a comprehensive set of experiments to carefully compare different models on multiple performance metrics. 

\subsubsection{Mixture of Gaussian}
We verify the conclusions in Proposition~\ref{prop:mixture_of_gaussian} by using the same setting in that proposition. We use a three layer deep network with 200 hidden units in each layer to simulate the highly flexible function family. For InfoVAE we choose the scaling coefficient $\lambda = 500$, information preference $\alpha=0$, and divergence optimized by MMD. 

\begin{figure*}[h]
\centering
\begin{tabular}{cc}
\includegraphics[width=0.4\linewidth]{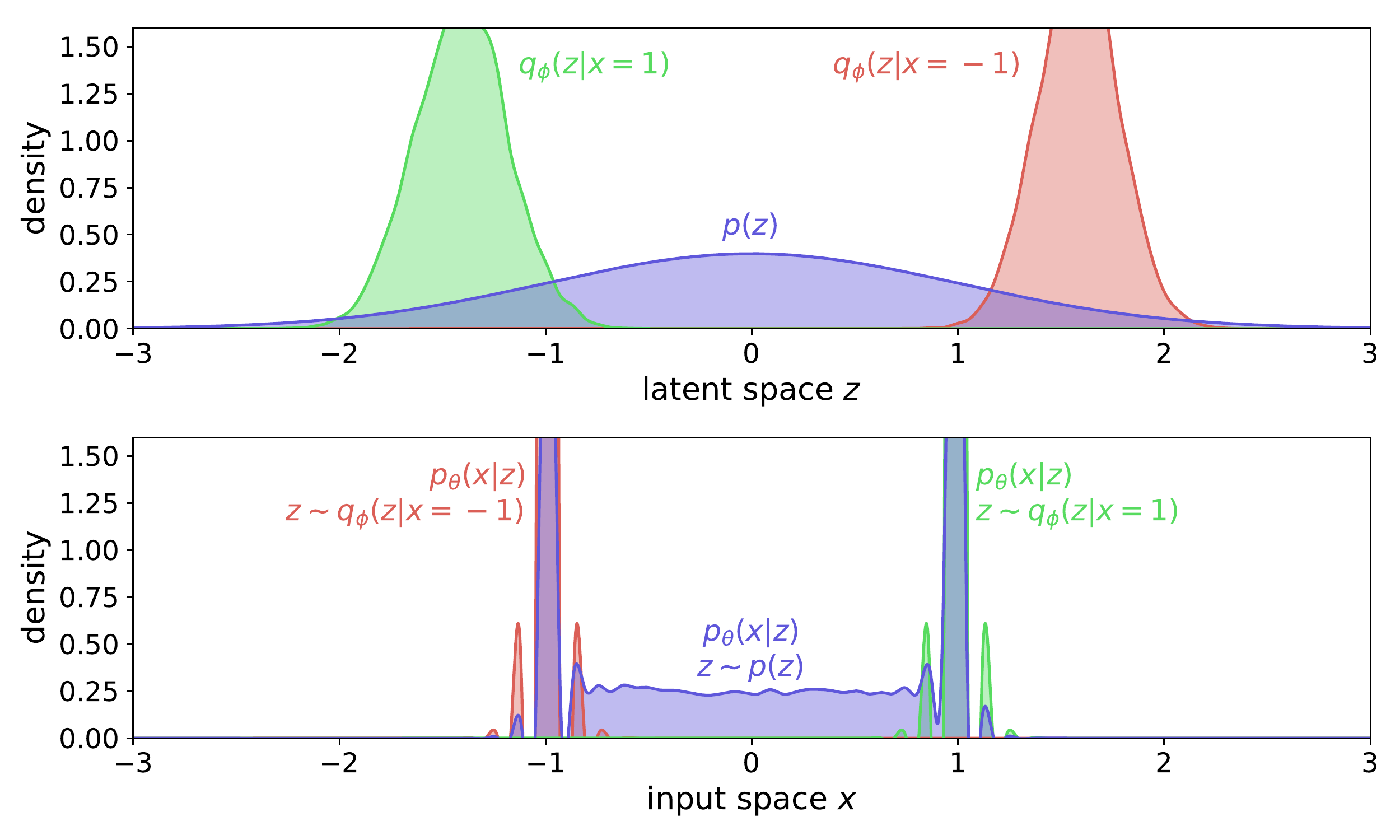} &
\includegraphics[width=0.4\linewidth]{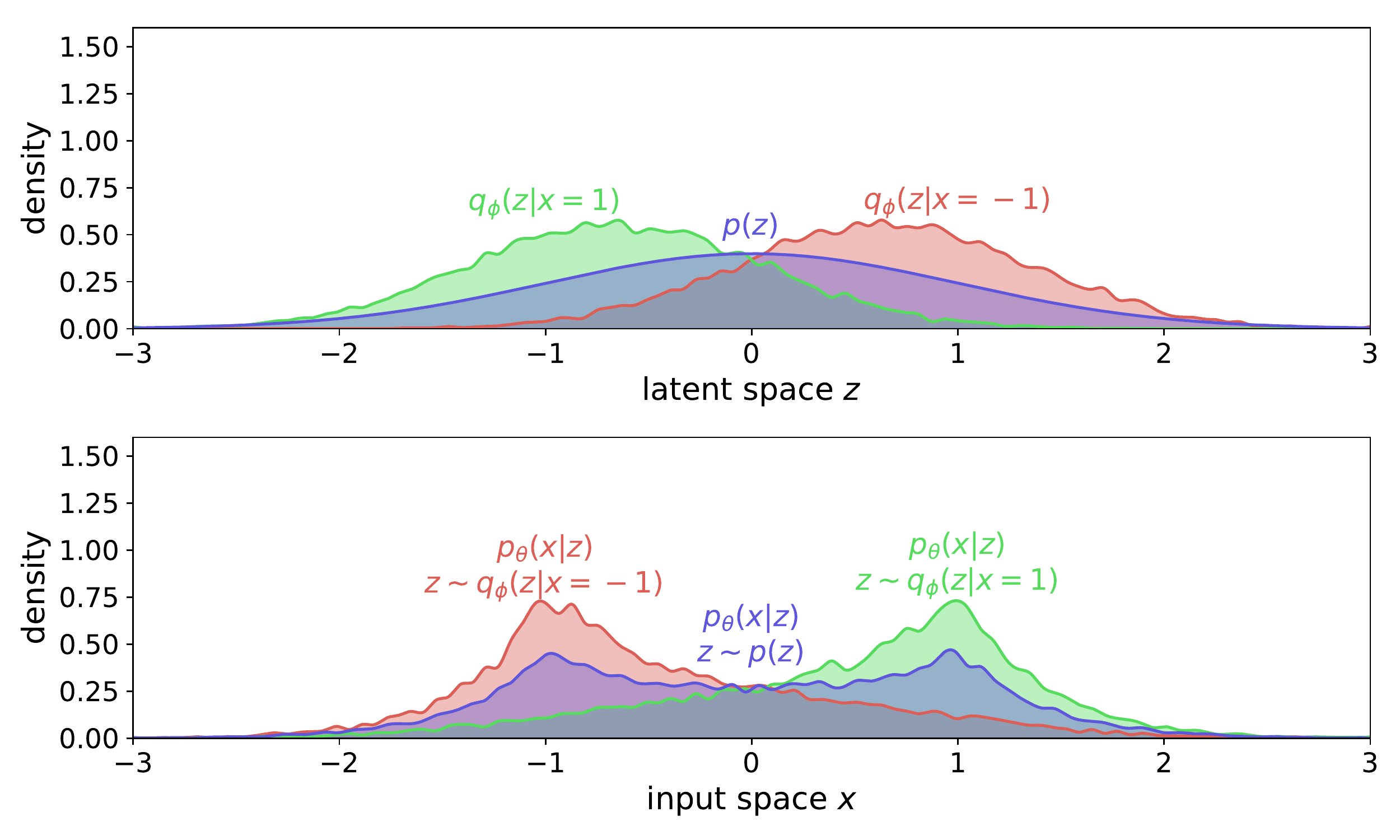} \\ 
\small{ELBO} & \small{InfoVAE ($\lambda=500$)}
\end{tabular}
\caption{Verification of Proposition~\ref{prop:mixture_of_gaussian} where the dataset only contains two examples $\lbrace -1, 1 \rbrace$. \textbf{Top:} density of the distributions $q_\phi(z|x)$ when $x=1$ (red) and $x=-1$ (green) compared with the true prior $p(z)$ (purple). \textbf{Bottom:} The  ``reconstruction'' $p_\theta(x|z)$ when $z$ is sampled from $q_\phi(z|x=1)$ (green) and $q_\phi(z|x=-1)$ (red). Also plotted is $p_\theta(x|z)$ when $z$ is sampled from the true prior $p(z)$ (purple). When the dataset consists of only two data points, ELBO (\textbf{left}) will push the density in latent space $Z$ away from $0$, while InfoVAE (\textbf{right}) does not suffer from this problem.}
\label{fig:mog_experiment}
\end{figure*}

The results are shown in Figure~\ref{fig:mog_experiment}. It can be observed that the predictions of the theory are reflected in the experiments: ELBO training leads to poor inference and a significantly over-estimated $q_\phi(z)$, while InfoVAE demonstrates a more stable behavior. 


\subsubsection{MNIST}
We demonstrate the problem on a real world dataset. We train ELBO and InfoVAE (with MMD regularization) on binarized MNIST with different training set sizes ranging from 500 to 50000 images; We use the DCGAN architecture \citep{deconvolutional_gan2015} for both models. For InfoVAE, we use the scaling coefficient $\lambda = 1000$, and information preference $\alpha=0$. We choose the number of latent features $\mathrm{dimension}(\mathcal{Z})=2$ to plot the latent space, and 10 for all other experiments. 

\begin{figure}[t]
\centering
\includegraphics[width=0.9\linewidth]{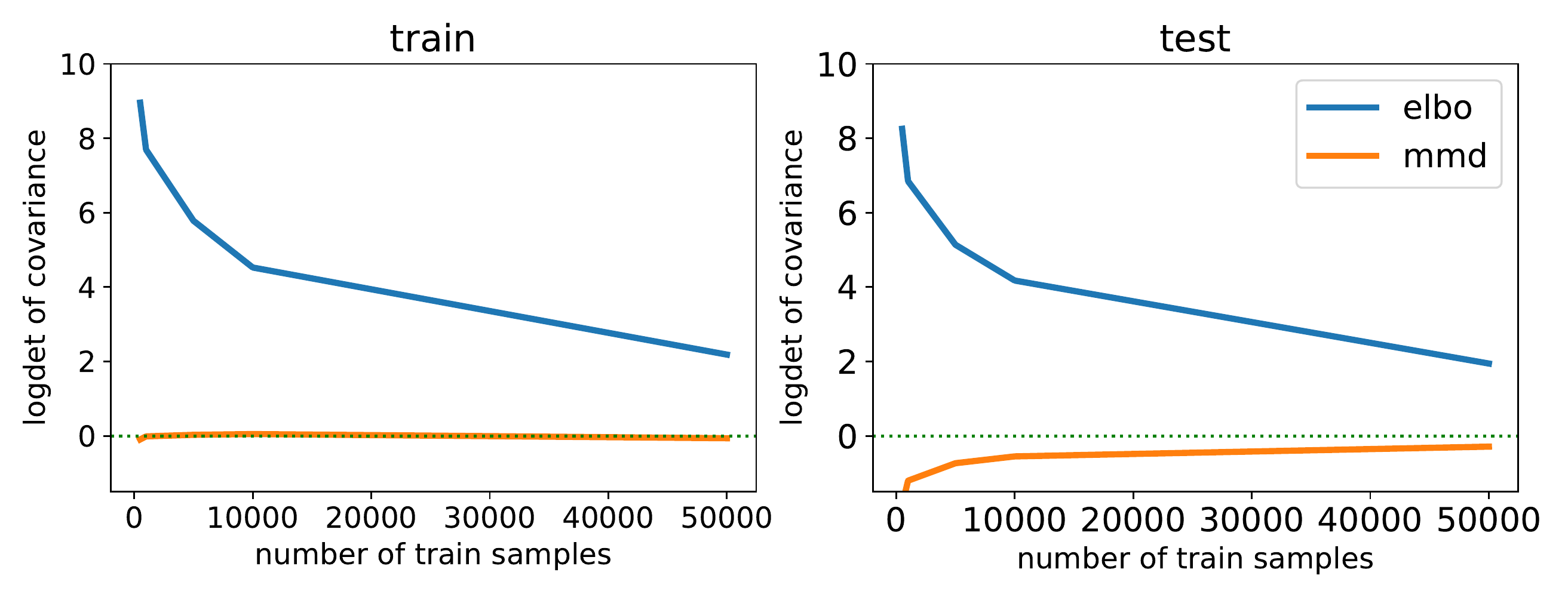} 
\vspace{-2ex}
\caption{$\log\det(\cov[q_\phi(z)])$ for ELBO vs. MMD-VAE under different training set sizes. The correct prior $p(z)$ has value $0$ on this metric, and values above or below $0$ correspond to over-estimation and under-estimation of the variance respectively. ELBO (blue curve) shows consistent over-estimation while InfoVAE does not. 
}
\label{fig:elbo_vs_mmd_latent}
\end{figure}

\begin{figure}[t]
\centering
\begin{tabular}{ccc}
\includegraphics[width=0.28\linewidth]{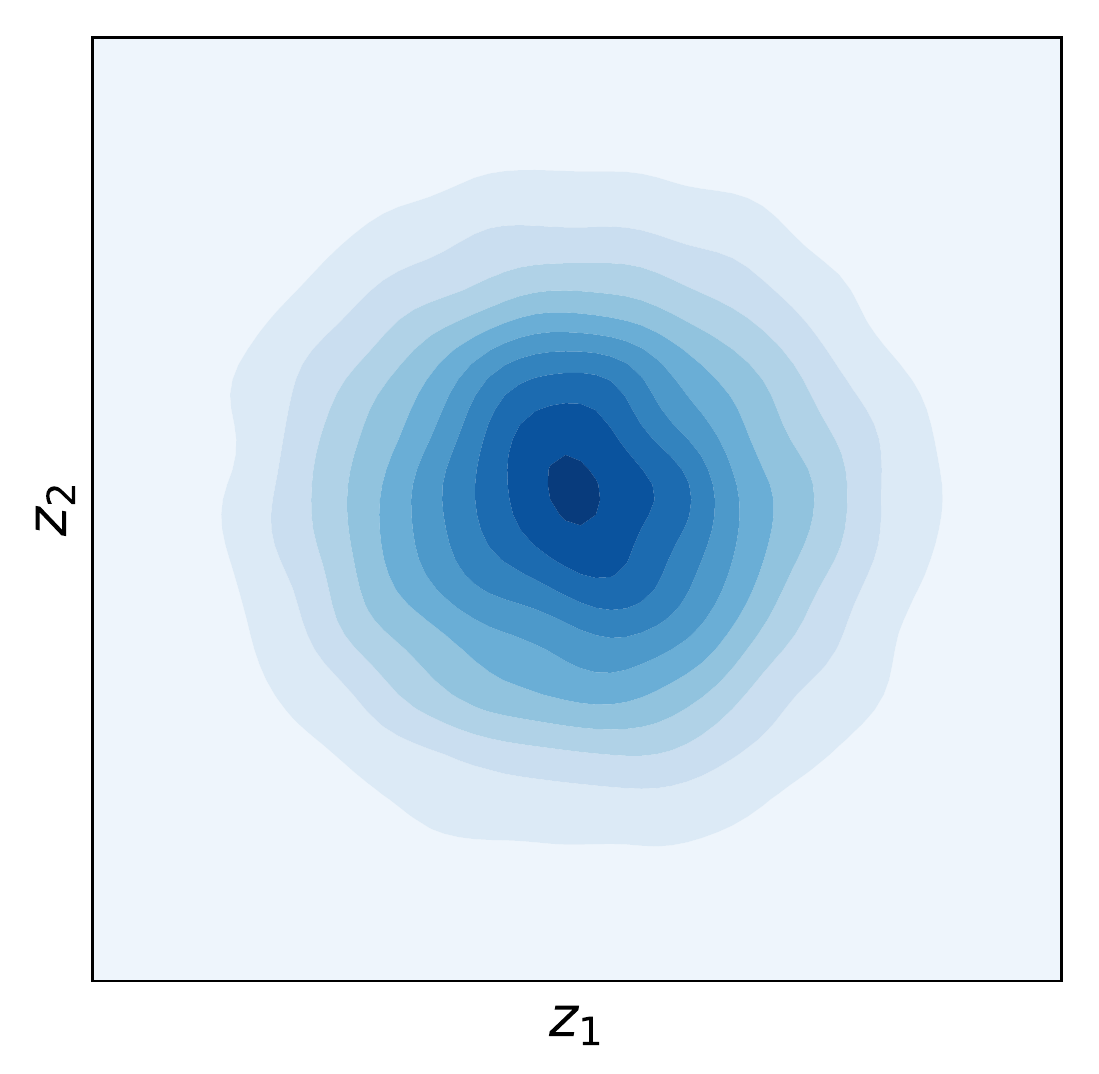} & 
\includegraphics[width=0.28\linewidth]{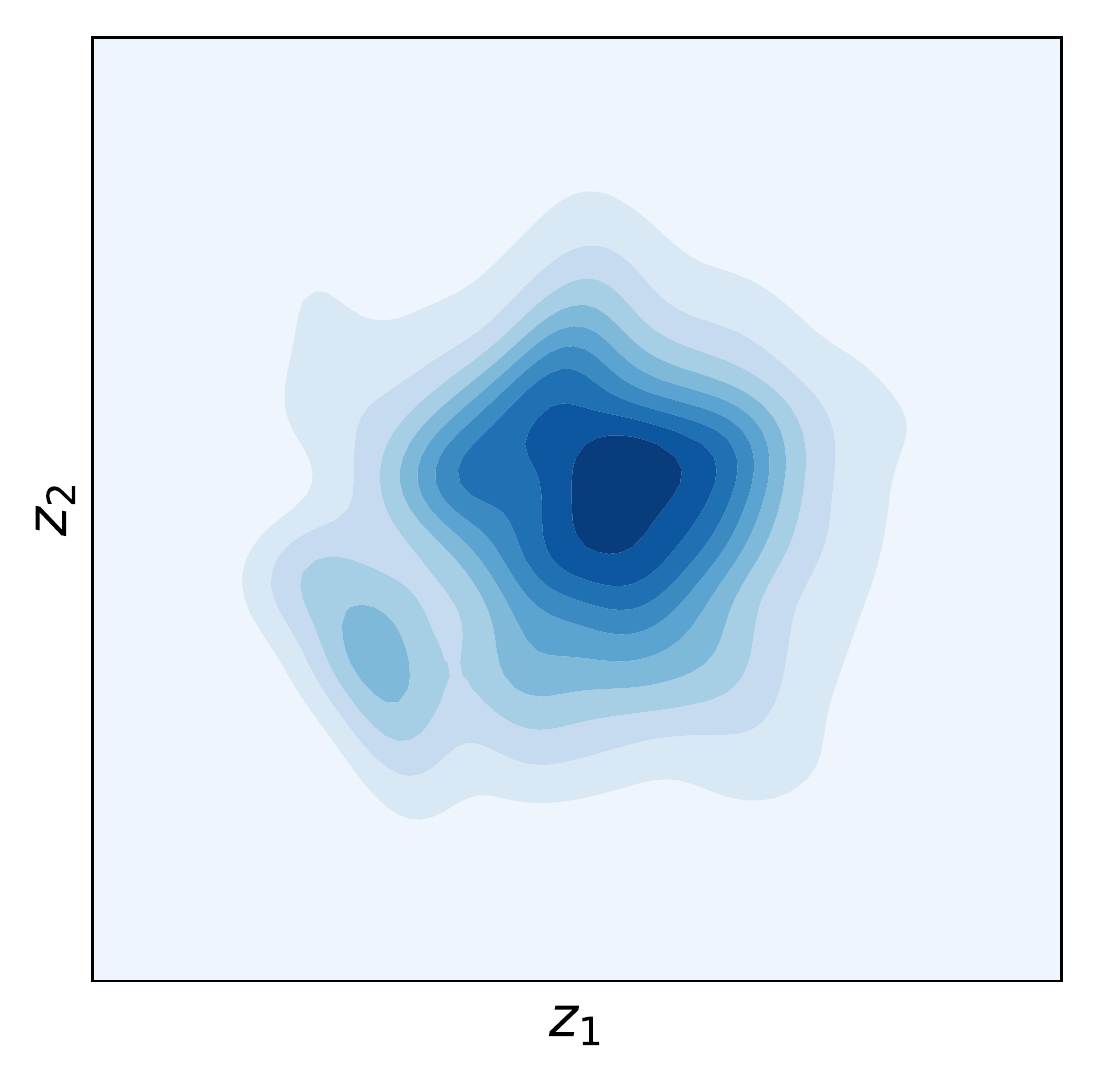} & 
\includegraphics[width=0.28\linewidth]{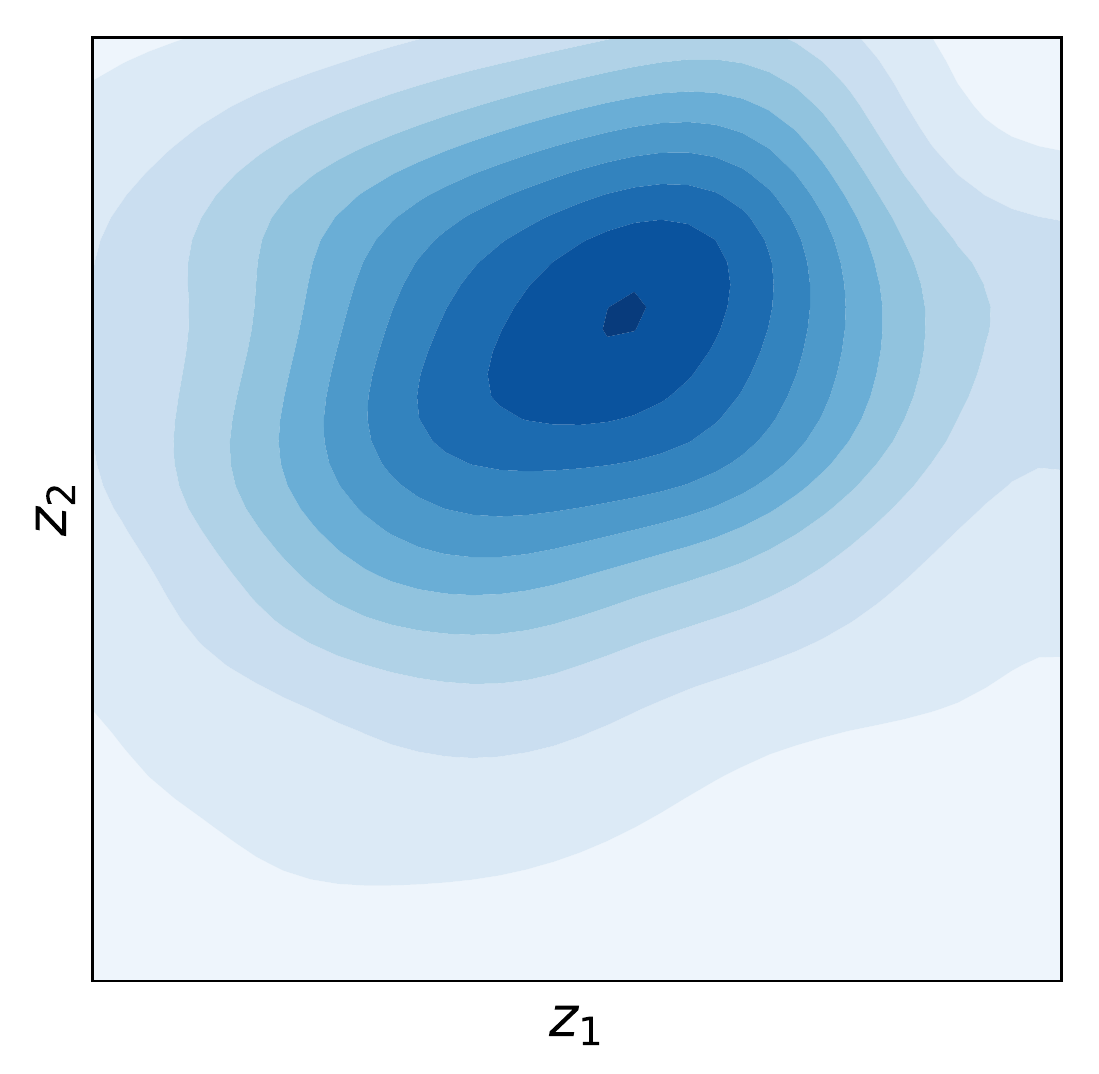} \\
Prior $p(z)$ & InfoVAE $q_\phi(z)$ & ELBO $q_\phi(z)$ 
\end{tabular}
\caption{Comparing the prior, InfoVAE $q_\phi(z)$ and ELBO $q_\phi(z)$. InfoVAE $q_\phi(z)$ is almost identical to the true prior, while the ELBO $q_\phi(z)$ is very far off. $q_\phi(z)$ for both models is computed by $\int_x p_\data q_\phi(z|x)$ where $p_\data$ is the test set.}
\label{fig:mnist_latent}
\end{figure}

\begin{figure}[t]
\centering
\begin{tabular}{ccc}
\includegraphics[width=0.45\linewidth]{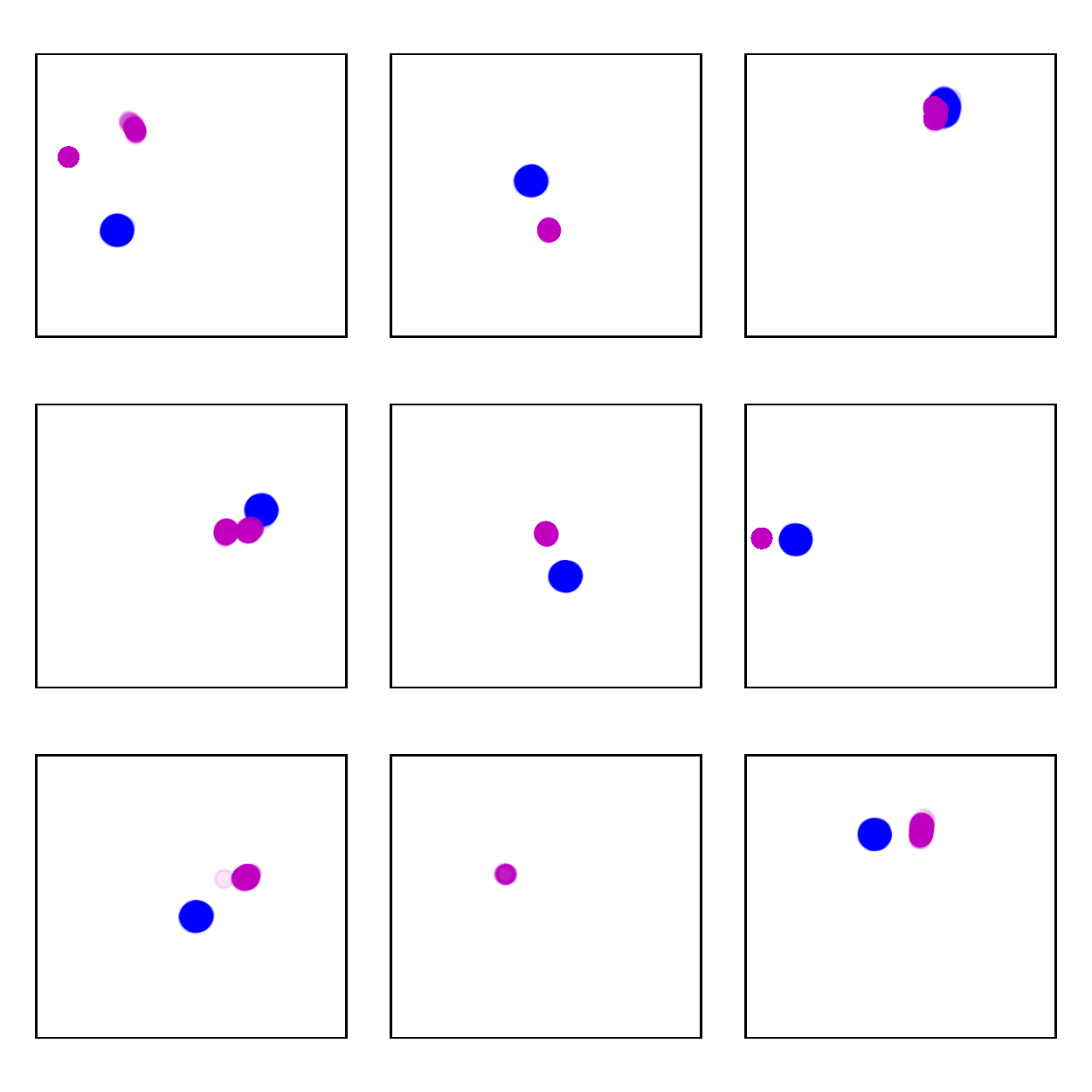} & 
\includegraphics[width=0.45\linewidth]{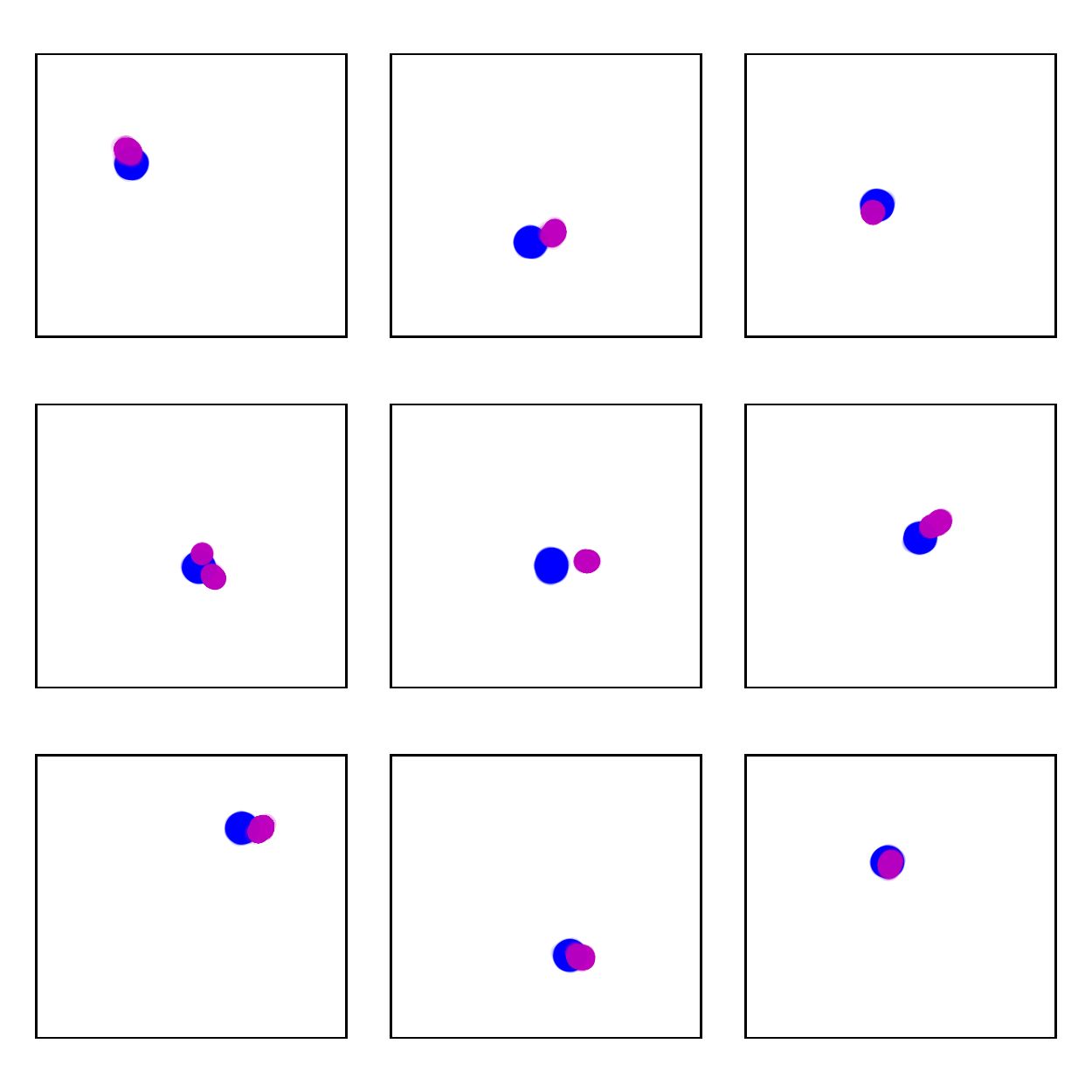} \\ 
ELBO Posterior & InfoVAE Posterior
\end{tabular}
\caption{Comparing true posterior $p_\theta(z|x)$ (green, generated by importance sampling) with approximate posterior $q_\phi(z|x)$ (blue) on testing data after training on 500 samples. For ELBO the approximate posterior is generally further from the true posterior compared to InfoVAE. (All plots are drawn with the same scale)}
\label{fig:mnist_posterior}
\end{figure}

First we verify that in real datasets ELBO over-estimates the variance of $q_\phi(z)$, while InfoVAE does not (with recommended choice of $\lambda$). In Figure~\ref{fig:elbo_vs_mmd_latent} we plot estimates for the log determinant of the covariance matrix of $q_\phi(z)$, denoted as $\log \det (\cov [q_\phi(z)])$ as a function of the size of the training set. For standard factored Gaussian prior $p(z)$, $\cov [p(z)] = I$, so $\log \det (\cov [p(z)]) = 0$. Values above or below zero give us an estimate of over or under-estimation of the variance of $q_\phi(z)$, which should in theory match the prior $p(z)$. It can be observed that the for ELBO, variance of $q_\phi(z)$ is significantly over-estimated. This is especially severe when the training set is small. On the other hand, when we use a large value for $\lambda$, InfoVAE can avoid this problem. 

To make this more intuitive we plot in Figure~\ref{fig:mnist_latent} the contour plot of $q_\phi(z)$ when training on 500 examples. It can be seen that with ELBO $q_\phi(z)$ matches $p(z)$ very poorly, while InfoVAE matches significantly better. 

To verify that ELBO trains inaccurate amortized inference we plot in Figure~\ref{fig:mnist_posterior} the comparison between samples from the approximate posterior $q_\phi(z|x)$ and samples from the true posterior $p_\theta(z|x)$ computed by rejection sampling.
The same trend can be observed. ELBO consistently gives very poor approximate posterior, while the InfoVAE posterior is mostly accurate.



\begin{figure}[t]
\centering
\begin{tabular}{cc}
\includegraphics[height=0.25\linewidth]{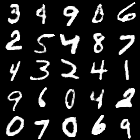} & 
\includegraphics[height=0.25\linewidth]{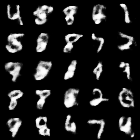} \\
\includegraphics[height=0.25\linewidth]{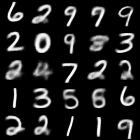} & 
\includegraphics[height=0.25\linewidth]{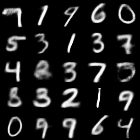} \\ 
reconstruction & generation
\end{tabular}
\caption{Samples generated by ELBO vs. MMD InfoVAE ($\lambda = 1000$) after training on 500 samples (plotting mean of $p_\theta(x|z)$). \textbf{Top:} Samples generated by ELBO. Even though ELBO generates very sharp reconstruction for samples on the training set, model samples $p(z)p_\theta(x|z)$ is very poor, and differ significantly from the reconstruction samples, indicating over-fitting, and mismatch between $q_\phi(z)$ and $p(z)$. \textbf{Bottom:} Samples generated by InfoVAE. The reconstructed samples and model samples look similar in quality and appearance, suggesting better generalization in the latent space.}
\label{fig:mnist_samples}
\end{figure}

Finally show the samples generated by the two models in Figure~\ref{fig:mnist_samples}. ELBO generates very sharp reconstructions, but very poor samples when sampled ancestrally $x \sim p(z)p_\theta(x|z)$. InfoVAE, on the other hand, generates samples of consistent quality, and in fact, produces samples of reasonable quality after only training on a dataset of 500 examples. This reflect InfoVAE's ability to control overfitting and demonstrate consistent training time and testing time behavior. 




\begin{figure*}[h]
\centering
\includegraphics[width=0.8\linewidth]{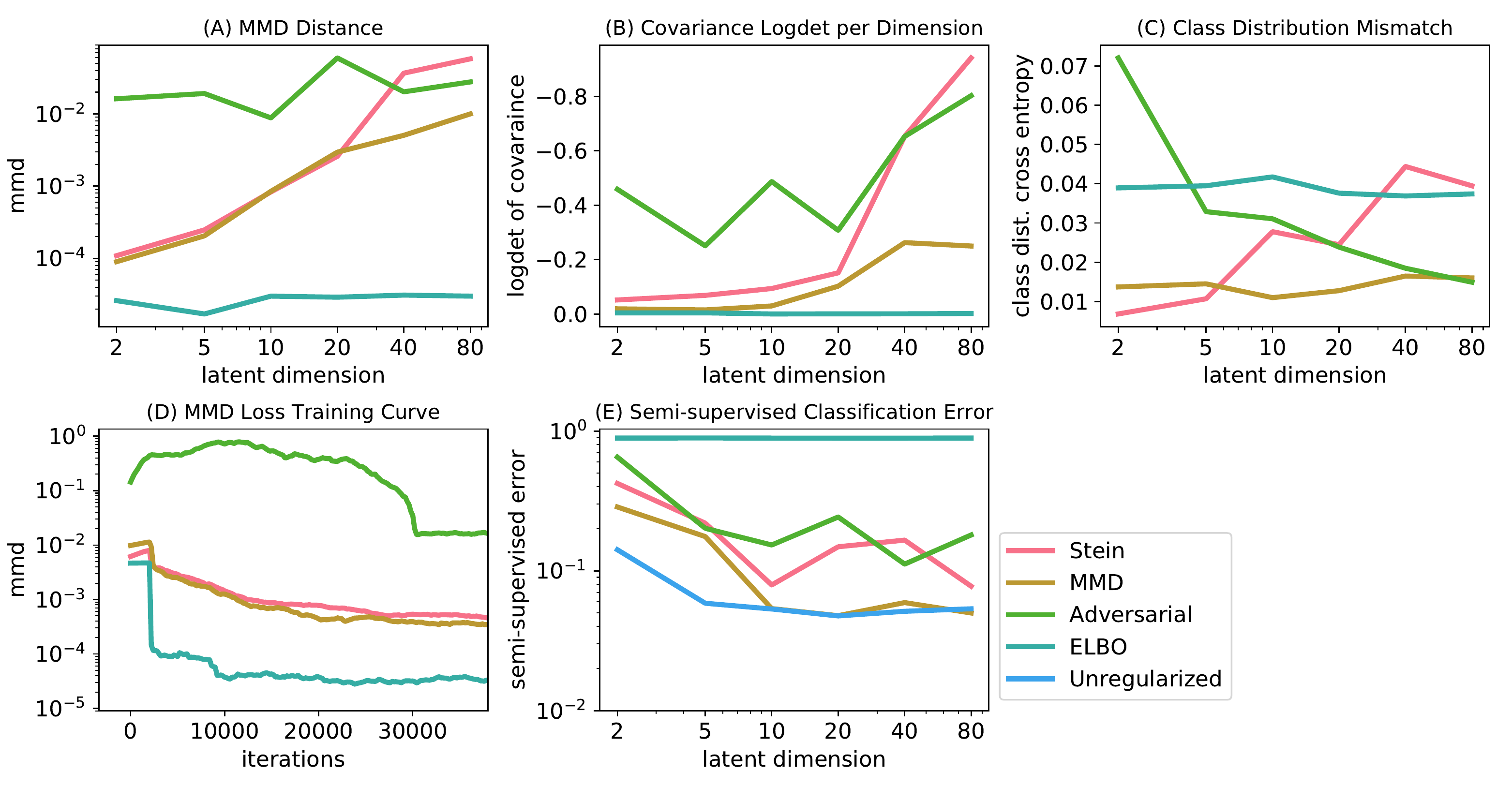}
\caption{Comparison of numerical performance. We evaluate MMD, log determinant of sample covariance, cross entropy with correct class distribution, and semi-supervised learning performance. `Stein', `MMD', `Adversarial' and `ELBO' corresponds to the VAE where the latent code are regularized with the respective methods, and `Unregularized' corresponds to the vanilla autoencoder without regularization over the latent dimensions.}
\label{fig:numerical_results}
\end{figure*}

\subsection{Comprehensive Comparison}
\label{sec:comprehensive_experiments}
In this section, we perform extensive qualitative and quantitative experiments on the binarized MNIST dataset to evaluate the performance of different models.  We would like to answer these questions:

\textbf{1)} Compare the models on a comprehensive set of numerical metrics of performance. Also compare the stability and training speed of different models.

\textbf{2)} Evaluate and compare the possible types of divergences (Adversarial, Stein, MMD).

For all two questions, we find InfoVAE with MMD regularization to perform better in almost all metrics of performance and demonstrate the best stability and training speed. The details are presented in the following sections. 

For models we use ELBO, Adversarial autoencoders, InfoVAE with Stein variational gradient, and InfoVAE with MMD ($\alpha=1$ because information preference is a concern, $\lambda=1000$ which can put the loss on $\mathcal{X}$ and $\mathcal{Z}$ on the same order of magnitude). In this setting we also use a highly flexible PixelCNN as the decoder $p_\theta(x|z)$ so that information preference is also a concern. Detailed experimental setup is explained in the Appendix. 



We consider multiple quantitative evaluations, including the quality of the samples generated, the training speed and stability, the use of latent features for semi-supervised learning, and log-likelihoods on samples from a separate test set.

\textbf{Distance between $q_\phi(z)$ and  $p(z)$:} 
To measure how well $q_\phi(z)$ approximates $p(z)$, we use two numerical metrics. The first is the full batch MMD statistic over the full data. Even though MMD is also used during training of MMD-VAE, it is too expensive to train using the full dataset, so we only use mini-batches for training. However during evaluation we can use the full dataset to obtain more accurate estimates. The second is the log determinant of the covariance matrix of $q_\phi(z)$. Ideally when $p(z)$ is the standard Gaussian $\Sigma_{q_\phi}$ should be the identity matrix, so $\log \det(\Sigma_{q_\phi}) = 0$. In our experiments we plot the log determinant divided by the dimensionality of the latent space. This measures the average under/over estimation per dimension of the learned covariance.

The results are plotted in Figure~\ref{fig:numerical_results} (A,B). This is different from the experiments in Figure~\ref{fig:elbo_vs_mmd_latent} because in this case the decoder is a highly complex pixel recurrent model and the concern that we highlight is failure to use latent features rather than inaccurate posterior. 
MMD achieves the best performance except for ELBO. Even though ELBO achieves extremely low error, this is trivial because for this experimental setting of flexible decoders, ELBO learns a latent code $z$ that does not contain any information about $x$, and $q_\phi(z|x) \approx p(z)$ for all $z$. 

\textbf{Sample distribution:} If the generative model $p(z)p_\theta(x|z)$ has true marginal $p_{data}(x)$, then the distribution of different object classes should also follow the distribution of classes in the original dataset. On the other hand, an incorrect generative model is unlikely to generate a class distribution that is identical to the ground truth. We let $\cv$ denote the class distribution in the real dataset, and $\hat{\cv}$ denote the class distribution of the generated images, computed by a pretrained classifier. We use cross entropy loss $l_{ce}(\cv, \hat{\cv}) = -\cv^T (\log \hat{\cv} - \log \cv)$ to measure the deviation from the true class distribution. 

The results for this metric are plotted in Figure~\ref{fig:numerical_results} (C). In general, Stein regularization performs well only with a small latent space with 2 dimensions, whereas the adversarial regularization performs better with larger dimensions; MMD regularization generally performs well in all the cases and the performance is stable with respect to the latent code dimensions.

\textbf{Training Speed and Stability:}
In general we would prefer a model that is stable, trains quickly and requires little hyperparameter tuning. In Figure~\ref{fig:numerical_results} (D) we plot the change of MMD statistic vs. the number of iterations. In this respect, adversarial autoencoder becomes less desirable because it takes much longer to converge, and sometimes converges to poor results even if we consider more power GAN training techniques such as Wasserstein GAN with gradient penalty~\cite{improved_wgan2017}. 

\textbf{Semi-supervised Learning:} 
To evaluate the quality of the learned features for other downstream tasks such as semi-supervised learning, we train a SVM directly on the learned latent features on MNIST images. We use the M1+TSVM in \citep{semi_supervised_by_variational_generative2014}, and use the semi-supervised performance over 1000 samples as an approximate metric to verify if informative and meaningful latent features are learned by the generative models. Lower classification error would suggest that the learned features $z$ contain more information about the data $x$. The results are shown in Figure~\ref{fig:numerical_results} (E). We observe that an unregularized autoencoder (which does not use any regularization $\mathcal{L}_{\mathrm{REG}}$) is superior when the latent dimension is low and MMD catches up when it is high. 
Furthermore, the latent code with the ELBO objective contains almost no information about the input and the semi-supervised learning error rate is no better than random guessing. 


\begin{table}
\centering
\caption{Log likelihood estimates for different models on the MNIST dataset. MMD-VAE achieves the best results, even though it is not explicitly optimizing a lower bound to the true log likelihood.}
\begin{tabular}{c|c}
\toprule
& Log likelihood estimate \\
\hline
ELBO & 82.75 \\
MMD-VAE & \textbf{80.76} \\
Stein-VAE & 81.47 \\
Adversarial VAE & 82.21 \\
\bottomrule 
 \end{tabular}
\label{table:log_likelihood}
\end{table}


\textbf{Log likelihood:} 
To be consistent with previous results, we use the stochastically binarized version first used in \citep{dbn_quantitative2008}. Estimation of log likelihood is achieved by importance sampling. We use 5-dimensional latent features in our log likelihood experiments. The values are shown in Table~\ref{table:log_likelihood}. Our results are slightly worse than reported in PixelRNN \citep{pixel_rnn2015}, which achieves a log likelihood of 79.2. However, all the regularizations perform on-par or superior compared to our ELBO baseline. This is somewhat surprising because we do not explicitly optimize a lower bound to the true log likelihood, unless we are using the ELBO objective. 

%% file: appendix.tex
\FloatBarrier
\newpage
\appendix
\section{Proofs}

\begin{proof}[Proof of Equivalence of ELBO Objectives]
\begin{align*}
\ELBO &= \Eb_{p_\data}\Eb_{q_\phi(z|x)}[\log p_\theta(x|z)] - \\
& \qquad \Eb_{p_\data}KL(q_\phi(z|x)\Vert p(z)) \\ 
&= \Eb_{q_\phi(x, z)}[\log p_\theta(x|z) + \log p(z) - \log q_\phi(z|x)] \\ 
&= \Eb_{q_\phi(x, z)}\left[\log \frac{p_\theta(x, z)}{q_\phi(x, z)} + \log p_\data(x) \right] \\
&= -\KL(q_\phi(x, z) \Vert p_\theta(x, z)) + E_{p_\data}[\log p_\data(x)] \\
\end{align*}
But $\Eb_{p_\data}[\log p_\data(x)]$ is a constant with no trainable parameters. 
\end{proof}

\begin{proof}[Proof of Equivalence of InfoVAE Objective Eq.(\ref{eq:info_vae}) to Eq.(\ref{eq:info_vae_tract})]
\begin{align*}
&\InfoVAE \\
&= -\lambda \KL(q_\phi(z) \Vert p(z)) - \\
& \qquad \Eb_{q(z)}[\KL(q_\phi(x|z) \Vert p_\theta(x|z)) ] + \alpha I_q(x; z) \\ 
&= \Eb_{q_\phi(x, z)}\left[ -\lambda \log \frac{q_\phi(z)}{p(z)} - \log \frac{q_\phi(x|z)}{p_\theta(x|z)} - \alpha \log \frac{q_\phi(z)}{q_\phi(z|x)} \right] \\
&= \Eb_{q_\phi(x, z)}\left[ \log p_\theta(x|z) - \log \frac{q_\phi(z)^{\lambda + \alpha - 1} p_\data(x)}{p(z)^\lambda q_\phi(z|x)^{\alpha-1}} \right] \\
&= \Eb_{q_\phi(x, z)}\left[ \log p_\theta(x|z) - \right. \\
& \qquad \qquad \left.\log \frac{q_\phi(z)^{\lambda + \alpha - 1} q_\phi(z|x)^{1 - \alpha} p_\data(x)}{p(z)^{\lambda+\alpha-1} p(z)^{1 - \alpha} } \right] \\
&= \Eb_{p_\data(x)} \Eb_{q_\phi(z|x)}[\log p_\theta(x|z)] - \\
& \qquad (1 - \alpha) \Eb_{p_\data(x)} \KL(q_\phi(z|x)||p(z)) - \\
& \qquad (\alpha + \lambda - 1) \KL(q_\phi(z) \Vert p(z)) - \Eb_{p_\data}[\log p_\data(x)]
\end{align*}
while the last term $\Eb_{p_\data}[\log p_\data(x)]$ is a constant with no trainable parameters.
\end{proof}

\begin{proof}[Proof of Proposition~\ref{prop:mixture_of_gaussian}]
Let the dataset contain two samples $\lbrace -1, 1 \rbrace$. Denote $\mathcal{L}_{\mathrm{AE}}(x)$ as $\Eb_{q_\phi(z|x)}[\log p_\theta(x|z)]$, and $\mathcal{L}_{\mathrm{REG}}(x)$ as $\KL(q_\phi(z|x) \Vert p(z))$. Due to the symmetricity in the problem, we constrict $p(x|z), q(z|x)$ to the following Gaussian distribution family: let $\sigma, c, \lambda \in \mathbb{R}$ be the parameters to optimize over, and
\begin{align*}
p(x|z) = \left\{
\begin{array}{ll}
\mathcal{N}(1, \sigma^2) & z \geq 0 \\ 
\mathcal{N}(-1, \sigma^2) & z < 0
\end{array}
\right. \\
q(z|x) = \left\{ 
\begin{array}{ll}
\mathcal{N}(c, \lambda^2) & x \geq 0 \\
\mathcal{N}(-c, \lambda^2) & x < 0 
\end{array}
\right.
\end{align*}
If $\ELBO$ can be maximized to $+\infty$ with this restricted family, then it can be maximized to $+\infty$ for arbitrary Gaussian conditional distributions. We have
\begin{align*}
\mathcal{L}_{\mathrm{AE}}&(x=1) \equiv E_{q(z|x=1)}[\log p(x=1|z)] \\ 
&= q(z \geq 0|x=1)\log p(x=1|z\geq0) + \\ 
&\qquad q(z < 0|x=1) \log p(x=1|z<0) \\ 
&= q(z \geq 0|x=1) (-1/2 \log(2\pi) - \log \sigma) + \\ 
& \qquad q(z < 0|x=1) (-1/2 \log(2\pi) - \log \sigma - 2/\sigma^2) \\ 
&= -\log \sigma - 2 q(z < 0|x=1) / \sigma^2 - 1/2\log(2\pi)
\end{align*}
The optimal solution for $\mathcal{L}_{\mathrm{AE}}(x=1)$ is achieved when
\begin{align*}
\frac{\partial \mathcal{L}_{\mathrm{AE}}(x=1)}{\partial \sigma} = -1/\sigma + 4/\sigma^3 q(z < 0|x=1) = 0
\end{align*}
where the unique valid solution is $\sigma = 2\sqrt{q(z < 0|x=1)}$, therefore optimally
\[ \mathcal{L}_{\mathrm{AE}}^*(x=1) = -1/2 \log q(z < 0|x=1) + C  \]
where $C \in \mathbb{R}$ is a constant independent of $c, \lambda$ and $\sigma$, and $q(z < 0|x=1)$ is the tail probability of a Gaussian. In the limit $\lambda \to 0, c \to \infty$, we have
\[ \mathcal{L}_{\mathrm{AE}}^*(x=1) = \Theta(c^2/\lambda^2) \]
Furthermore we have
\begin{align*}
\mathcal{L}_{\mathrm{REG}}(x=1) &= -\KL(q(z|x=1)||p(z)) \\
& = \log \lambda - \lambda^2/2 - c^2/2 + 1/2
\end{align*}
In addition because $\mathcal{L}_{\mathrm{REG}}$ has no dependence on $\sigma$, so the $\sigma$ that maximizes $\mathcal{L}_{\mathrm{AE}}$ also maximizes $\ELBO$. Therefore in the limit of $\lambda \to 0$, $c \to \infty$, and $\sigma$ chosen optimally, we have
\begin{align*}
\lim_{\lambda \to 0, c \to \infty} & \mathcal{L}_{\mathrm{ELBO}}(x=1)  \\
&= \mathcal{L}^*_{\mathrm{AE}}(x=1) + \mathcal{L}_{\mathrm{REG}}(x=1) \\
&= \Theta( c^2/\lambda^2 + \log \lambda - \lambda^2/2 - c^2/2 ) \\ 
&\to +\infty
\end{align*}
This means that the growth rate of the $\mathcal{L}_{\mathrm{ELBO}}(x=1)$ far exceeds the growth rate of $\mathcal{L}_{\mathrm{REG}}(x=1)$, and their sum is still maximized to $+\infty$. We can obtain similar conclusions for the symmetric case of $x = -1$. Therefore over-fitting to $\mathcal{L}_{ELBO}$ has unbounded reward when $c \to \infty$ and $\lambda \to 0$. 

To prove that the variational gap tends to $+\infty$, observe that when $x = 1$, for any $z \in \mathbb{R}$, 
\begin{align*}
p(z|x=1) &= \frac{p(x=1|z)}{p(x=1)}p(z) \\
&= \frac{p(x=1|z)}{\int_{z'} p(x=1|z') p(z') dz'} p(z) \\ 
&= \frac{p(x=1|z)}{1/2 p(x=1|z\geq 0) + 1/2 p(x=1|z < 0)} p(z) \\
&\leq \frac{2 p(x=1|z)}{p(x=1|z\geq 0)} p(z) \leq 2 p(z)
\end{align*}
This means that the variational gap \begin{align*}
\KL( & q(z|x=1) \Vert p(z|x=1)) \\
&= E_{q(z|x=1)}\left[ \log \frac{q(z|x=1)}{p(z|x=1)} \right] \\
&\geq E_{q(z|x=1)}\left[ \log \frac{q(z|x=1)}{2p(z)} \right] \\
&= \KL(q(z|x=1) \Vert p(z)) - \log 2 \\
&\to \infty 
\end{align*}
The same argument holds for $x=-1$.
\end{proof}


\begin{proof}[Proof of Proposition~\ref{prop:mixture_of_gaussian}]
Let the dataset contain two samples $\lbrace -1, 1 \rbrace$, and $p(x|z), q(z|x)$ be arbitrary one dimensional Gaussians, then by symmetry of the problem, at optimality for $\mathcal{L}_{ELBO}$, we have
\begin{align*}
p(x|z) = \left\{
\begin{array}{ll}
\mathcal{N}(1, \sigma^2) & z > 0 \\ 
\mathcal{N}(-1, \sigma^2) & z \leq 0
\end{array}
\right. \\
q(z|x) = \left\{ 
\begin{array}{ll}
\mathcal{N}(c, \lambda^2) & x > 0 \\
\mathcal{N}(-c, \lambda^2) & x \leq 0 
\end{array}
\right.
\end{align*}
Then we have
\begin{align*}
\mathcal{L}_{\mathrm{LL}}&(x=1) \equiv E_{q(z|x=1)}[\log p(x=1|z)] \\ 
&= q(z \geq 0|x=1)\log p(x=1|z\geq0) + \\ 
&\qquad q(z < 0|x=1) \log p(x=1|z<0) \\ 
&= q(z \geq 0|x=1) (-1/2 \log(2\pi) - \log \sigma) + \\ 
& \qquad q(z < 0|x=1) (-1/2 \log(2\pi) - \log \sigma - 2/\sigma^2) \\ 
&= -\log \sigma - 2 q(z < 0|x=1) / \sigma^2 - 1/2\log(2\pi)
\end{align*}
The optimal solution $\sigma^*$ is achieved where
\begin{align*}
\frac{\partial \mathcal{L}_{\mathrm{LL}}}{\partial \sigma} = -1/\sigma + 4/\sigma^3 = 0
\end{align*}
where the unique valid solution is $\sigma = 2\sqrt{q(z < 0|x=1)}$, therefore optimally
\[ \mathcal{L}_{\mathrm{LL}}^*(x=1) = -1/2 \log q(z < 0|x=1) - \log 2 - 1/2\log(2\pi)  \]
Where $q(z < 0|x=1)$ is the tail probability of a Gaussian. In the limit $\lambda \to 0, c \to \infty$, we have
\[ \mathcal{L}_{\mathrm{LL}}^*(x=1) = \Theta(c^2/\lambda^2) \]
Furthermore we have
\begin{align*}
\KL(q(z|x=1)||p(z)) = -\log \lambda + \lambda^2/2 + c^2/2 - 1/2
\end{align*}
Then in the limit of $\lambda \to 0$, $c \to \infty$
\begin{align*}
\lim_{\lambda \to 0, c \to \infty} & \mathcal{L}_{\mathrm{ELBO}}(x=1)  \\
&= - \mathcal{L}_{\mathrm{LL}}(x=1) + \KL(q(z|x=1)||p(z)) \\
&= \Theta( -c^2/\lambda^2 - \log \lambda + \lambda^2/2 + c^2/2 ) \\ 
&\to -\infty
\end{align*}
We can obtain similar conclusions for the symmetric case of $x = -1$. Therefore over-fitting to $\mathcal{L}_{ELBO}$ has unbounded reward when $c \to \infty$ and $\lambda \to 0$. In addition, because $c \to \infty$, but the true posterior $p(z|x)$ has bounded second order moments, the variational gap $\KL(q(z|x)\Vert p(z|x)) \to \infty$.
\end{proof}

\begin{proof}[Proof of Proposition~\ref{prop:infovae_correctness}]
We first rewrite the modified $\InfoVAEh$ objective 
\begin{align*}
\InfoVAEh &= \Eb_{q_\phi(z, x)}[\log p_\theta(x|z)]  \\
& \qquad -(1 - \alpha) \Eb_{p_{\data}}[\KL(q(z|x)\Vert p(z))]  \\ 
& \qquad -(\alpha + \lambda - 1) D(q_\phi(z)||p(z)) \\
&= \Eb_{q_\phi(z, x)}[\log p_\theta(x|z)]  - (1 - \alpha) I_{q_\phi}(x; z)  \\
& \qquad -(1 - \alpha) \KL(q_\phi(z) \Vert p(z))  \\
& \qquad -(\alpha + \lambda - 1) D(q_\phi(z)||p(z)) \\
& \qquad 
\end{align*}
Notice that by our condition of $\alpha < 1, \lambda > 0$, we have $1 - \alpha > 0, \alpha + \lambda - 1 > 0$. For convenience we will rename 
\begin{align*}
\beta &\equiv 1 - \alpha > 0 \\
\gamma &\equiv \alpha + \lambda - 1 > 0
\end{align*}
In addition we consider our objective in two separate terms
\begin{align*}
\mathcal{L}_1 &= \Eb_{q_\phi(z, x)}[\log p_\theta(x|z)]  - \beta I_{q_\phi}(x; z) \\
\mathcal{L}_2 &= -\beta \KL(q_\phi(z) \Vert p(z)) - \gamma D(q_\phi(z)||p(z))
\end{align*}

We will prove that whenever $\beta > 0, \gamma > 0$, the two terms are maximized under the condition in the proposition respectively. First consider $\mathcal{L}_1$, because $I_{q_\phi}(x, z) = I_0$:
\begin{align*}
\mathcal{L}_1 &= \Eb_{q_\phi(z, x)}[\log p_\theta(x|z)] - \beta I_0 \\
&= \Eb_{q_\phi(z)} \Eb_{q_\phi(x|z)}[\log p_\theta(x|z)] - \beta I_0\\ 
&= \Eb_{q_\phi(z)}\left[ -\KL(q_\phi(x|z)||p_\theta(x|z)) + H_{q_\phi}(x|z) \right] \\
& \qquad - \beta I_0
\end{align*}
Therefore for any $q_\phi(z|x)$, 
$p_{\theta^*}(x|z)$ that optimizes $\mathcal{L}_1$ satisfies $\forall z$, $p_{\theta^*}(x|z) = q_\phi(x|z)$, and we have for any given $q_\phi$, the optimal $\mathcal{L}_1$ is 
\begin{align*}
\mathcal{L}^*_1 &= \E_{q_\phi(x, z)}[\log p_{\theta^*}(x|z)] - \beta I_0 \\
&= \E_{q_\phi(z)}\E_{q_\phi(x|z)} [\log q_\phi(x|z)] - \beta I_0 \\
&= \E_{q_\phi(z)}[-H_{q_\phi}(x|z)] - \beta I_0 \\
&= (1 - \beta) I_0 - H_{p_{\data}}(x) \label{eq:mi}
\end{align*}
where we use $H_p(x)$ to denote the entropy of $p(x)$. Notice that $\mathcal{L}_1$ is dependent on $q_\phi$ only by $I_{q_\phi}(x; z)$ therefore when $I_{q_\phi}(x; z) = I_0$, $\mathcal{L}_1$ is maximized regardless of the choice of $q_\phi$. So we only have to independently maximize $\mathcal{L}_2$ subject to fixed $I_0$. Notice that $\mathcal{L}_1$ is maximized when $q_\phi(z) = p(z)$, and we show that this can be achieved. When $\lbrace q_\phi \rbrace$ is sufficiently flexible we simply have to partition the support set $\mathcal{A}$ of $p(z)$ into $N=\lceil e^{I_0} \rceil$ subsets $\lbrace \mathcal{A}_1, \cdots, \mathcal{A}_N \rbrace$, so that each subset satisfies $\int_{\mathcal{A}_i} p(z) dz = 1/N$. Similarly we partition the support set $\mathcal{B}$ of $p_{\data}(x)$ into $N$ subsets $\lbrace \mathcal{B}_1, \cdots, \mathcal{B}_N \rbrace$, so that each subset satisfies $\int_{\mathcal{B}_i} p_{\data}(x) dx = 1/N$. Then we construct $q_\phi(z|x)$ mapping each $B_i$ to $A_i$ as follows
\begin{align*}
q(z|x) = \left\lbrace \begin{array}{ll} N p(z) & z \in A_i \\ 0 & \mathrm{otherwise} \end{array}
\right.
\end{align*} 
such that for any $x_i \in B_i$. It is easy to see that this distribution is normalized because
\[ \int_z q(z|x) dz = \int_{\mathcal{A}_i} Np(z) dz = 1\]
Also it is easy to see that $p(z) = q_\phi(z)$. In addition
\begin{align*}
&I_q(x; z) = H_q(z) - H_q(z|x) \\
&= H_q(z) + \int_\mathcal{B} p_{\data}(x) \int_{\mathcal{A}} q(z|x) \log q(z|x) dz dx \\
&= H_q(z) + \frac{1}{N} \sum_i \int_{\mathcal{B}_i} \int_{\mathcal{A}} q(z|x) \log q(z|x) dz dx \\
&= H_q(z) + \frac{1}{N} \sum_i \int_{\mathcal{A}_i} N q(z) \log (N q(z)) dz \\ 
&= H_q(z) + \sum_i \int_{\mathcal{A}_i} q(z) \left(I_0 + \log q(z)\right) dz \\
&= H_q(z) + \int_{\mathcal{A}} q(z) \left(I_0 + \log q(z)\right) dz \\
&= H_q(z) + I_0 - H_q(z) = I_0
\end{align*}
Therefore we have constructed a $q_\phi(z|x), p_\theta(x|z)$ so that we have reached the maximum for both objectives 
\begin{align*}
&\mathcal{L}_1^* = (1 - \beta) I_0 - H_{p_{\data}}(x) \\
&\mathcal{L}_2^* = 0
\end{align*}
so there sum must also be maximized. Under this optimal solution we have that $q_\phi(x|z) = p_\theta(x|z)$ and $q_\phi(z) = p(z)$, this implies $q_\phi(x, z) = p_\theta(x, z)$, which implies that both $p_\theta(z|x) = q_\phi(z|x)$ and $p_\theta(x) = p_{\data}(x)$.
\end{proof}

\section{Stein Variational Gradient}
The Stein variational gradient \citep{stein_variational2016} is a simple and effective framework for matching a distribution $q$ to $p$ by descending the variational gradient of $\KL(q(z)||p(z))$. Let $q(z)$ be some distribution on $\mathcal{Z}$, $\epsilon$ be a small step size, $k$ be a positive definite kernel function, and $\phi(z)$ be a function $\mathcal{Z} \to \mathcal{Z}$. Then $T(z) = z + \epsilon \phi(z)$ defines a transformation, and this transformation induces a new distribution $q_{[T]}(z')$ on $\mathcal{Z}$ where $z' = T(z), z \sim q(z)$. Then the $\phi^*$ that minimizes $\KL(q_{[T]}(z)||p(z))$, as $\epsilon \to 0$ is
\[ \phi^*_{q, p}(\cdot) = \Eb_{z \sim q} [k(z, \cdot)\nabla_z \log p(z) + \nabla_z k(z, \cdot)] \]
as shown in Lemma 3.2 in \citep{stein_variational2016}.
Intuitively $\phi^*_{q, p}$ is the steepest direction that transforms $q(z)$ towards $p(z)$. In practice, $q(z)$ can be represented by the particles in a mini-batch.

We propose to use Stein variational gradient to regularize variational autoencoders using the following process. For a mini-batch $\xv$, we compute the corresponding mini-batch features $\zv = q_\phi(\xv)$. Based on this mini-batch we compute the Stein gradients by empirical samples
\[ \phi^*_{q_\phi, p}(z_i) \approx \frac{1}{n} \sum_{j=1}^{n} k(z_i, z_j) \nabla_{z_j} \log p(z_j) + \nabla_{z_j} k(z_j, z_i) \]
The gradients wrt. the model parameters are
\[ \nabla_{\phi} \KL(q_\phi(z) || p(z)) \approx \frac{1}{n} \sum_{i=1}^{n} \nabla_\phi q_\phi(z_i)^T \phi^*_{q_\phi, p}(z_i) \]
In practice we can define a surrogate loss
\[ \hat{D}(q_\phi(z), p(z)) = z^T \mbox{stop\_gradient}(\phi^*_{q_\phi, p} (z)) \]
where $\mbox{stop\_gradient}(\cdot)$ indicates that this term is treated as a constant during back propagation. Note that this is not really a divergence, but simply a convenient loss function that we can implement using standard automatic differentiation software, whose gradient is the Stein variational gradient of the true KL divergence. 

\section{Experimental Setup}
In all our experiments in Section~\ref{sec:comprehensive_experiments}, we choose the prior $p(z)$ to be a Gaussian with zero mean and identity covariance, $p_\theta(x|z)$ to be a PixelCNN conditioned on the latent code \citep{conditional_pixelcnn2016}, and $q_\phi(z|x)$ to be a CNN where the final layer outputs the mean and standard deviation of a factored Gaussian distribution. 

For MNIST we use a simplified version of the conditional PixelCNN architecture \citep{conditional_pixelcnn2016}. For CIFAR we use the public implementation of PixelCNN++ \citep{pixelcnn_pp2017}. In either case we use a convolutional encoder with the same architecture as \citep{deconvolutional_gan2015} to generate the latent code, and plug this into the conditional input for both models. The entire model is trained end to end by Adam \citep{adam_optimization2014}. PixelCNN on ImageNet take about 10h to train to convergence on a single Titan X, and CIFAR take about two days to train to convergence. We will make the code public upon publication.


\section{CIFAR Samples}
\begin{figure*}[h]
\centering
\begin{tabular}{cccccc}
\includegraphics[height=0.13\linewidth]{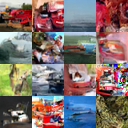} & 
\includegraphics[height=0.13\linewidth]{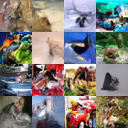} & 
\includegraphics[height=0.13\linewidth]{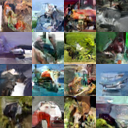} &
\includegraphics[height=0.13\linewidth]{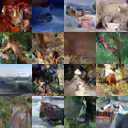} &
\includegraphics[height=0.13\linewidth]{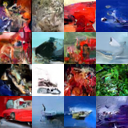} & 
\includegraphics[height=0.13\linewidth]{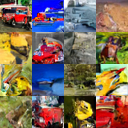} \\ 
ELBO-2 & ELBO-10 & MMD-2 & MMD-10 & Stein-2 & Stein-10 
\end{tabular}
\caption{CIFAR samples. We plots samples for ELBO, MMD, and Stein regularization, with 2 or 10 dimensional latent features. Samples generated with Stein regularization and MMD regularization are more coherent globally than those generated with ELBO regularization.}
\label{fig:cifar_samples}
\end{figure*}

We additional perform experiments on the CIFAR \citep{cifar2009} dataset. We show samples from models with different regularization methods - ELBO, MMD, Stein and Adversarial in Figure~\ref{fig:cifar_samples}. In all cases, the model accurately matches $q_\phi(z)$ with $p(z)$, and samples generated with Stein regularization and MMD regularization are more coherent globally compared to the samples generated with ELBO regularization.

